\documentclass[runningheads]{llncs}
\usepackage{graphicx}

\usepackage{multirow}
\usepackage{graphicx}
\usepackage{amsmath,amssymb,bm}

\usepackage{xcolor}
\usepackage[colorlinks=true,citecolor=blue,urlcolor=blue]{hyperref}

\begin{document}

\title{Robust Multimodal Brain Tumor Segmentation \\ via Feature Disentanglement and Gated Fusion}

\titlerunning{Robust Multimodal Brain Tumor Segmentation}
%

\author{Cheng Chen\inst{1}, Qi Dou\inst{2}, Yueming Jin\inst{1}, Hao Chen\inst{1,3}, \\Jing Qin\inst{4}, and Pheng-Ann Heng\inst{1,5}}
\institute{Dept. of Computer Science and Engineering, The Chinese University of Hong Kong
	\and Department of Computing, Imperial College Longdon
	\and Imsight Medical Technology Co. Ltd., China
	\and Centre for Smart Health, School of Nursing, The Hong Kong Polytechnic University
	\and Guangdong Provincial Key Laboratory of Computer Vision and Virtual Reality Technology, SIAT, Chinese Academy of Sciences, China}
\authorrunning{C. Chen et al.}
%
%
\maketitle              
\begin{abstract}
Accurate medical image segmentation commonly requires effective learning of the complementary information from multimodal data.
However, in clinical practice, we often encounter the problem of missing imaging modalities. 
We tackle this challenge and propose a novel multimodal segmentation framework which is robust to the absence of imaging modalities. 
Our network uses feature disentanglement to decompose the input modalities into the modality-specific appearance code, which uniquely sticks to each modality, and the modality-invariant content code, which 
absorbs multimodal information for the segmentation task. 
With enhanced modality-invariance, the disentangled content code from each modality is fused into a shared representation which gains robustness to missing data. 
The fusion is achieved via a learning-based strategy to gate the contribution of different modalities at different locations. 
We validate our method on the important yet challenging multimodal brain tumor segmentation task with the BRATS challenge dataset. With competitive performance to the state-of-the-art approaches for full modality, our method achieves outstanding robustness under various missing modality(ies) situations, 
significantly exceeding the state-of-the-art method by over $16\%$ in average for Dice on whole tumor segmentation.

\end{abstract}

\section{Introduction}
Accurate segmentation of brain tumor is of critical importance for quantitative assessment of tumor progression and preoperative treatment planning. 
The measurement of tumor-induced tissue changes relies on complementary biological information provided in multiple Magnetic Resonance Imaging (MRI) modalities, i.e., FLAIR, T1, T1 contrast-enhanced (T1c), and T2.
Joint learning from these multimodal images greatly helps to improve segmentation accuracy.
A plentiful of multimodal methods have been developed for automated brain tumor segmentation, by either concatenating multiple MRI modalities as inputs~\cite{kamnitsas2017efficient,zhou2018one}, or fusing higher-level features from each modality in latent space~\cite{fidon2017scalable,tseng2017joint}.
However, availability of a full set of the desired modalities is not always guaranteed in real-world scenarios, due to various scanning protocols and diverse patient conditions.
In this regard, robustness to one or more missing modality(ies) during inference is essential for a widely-applicable multimodal learning method.

A typical solution is to synthesize the missing modality(ies) with available ones~\cite{van2015does}.
Such method requires to build a specific model for each modality from all possible combinations of available modalities, which is complicated.
Alternatively, Havaei et al.~\cite{havaei2016hemis} propose hetero-modal image segmentation (HeMIS), which fuses multimodal information by computing statistics (i.e., mean and variance) across individual features.
This method is easily scalable to various data missing situations, as the fusion in latent space adapts to any number of modalities.
Furthermore, Chartsias et al.~\cite{chartsias2018multimodal} and Van Tulder et al.~\cite{van2019learning} enhance the modality-invariance of latent representations by minimizing the 
L1 or L2 distance of features from different modalities.
However, different MRI modalities vary in intensity distributions with modality-specific appearance, thus simply encouraging the features from different modalities to be close under L1- or L2-Norm may not achieve optimal representations with desired modality-invariance. 
Instead, a concurrent work~\cite{shen2019brain} uses adversarial learning to ensure the model generate similar features under missing modalities as in a full modality situation.

To effectively extract modality-invariant representations conveying essential content of tumor, learning to cancel out the modality-specific information may be helpful.
This can be achieved using feature disentanglement by decomposing inputs to latent space of interpretable factors~\cite{huang2018multimodal,tsai2018learning}.
In medical imaging, disentangling representations has recently demonstrated effectiveness for liver lesion classification~\cite{ben2018improving}, myocardial segmentation~\cite{chartsias2018factorised} and multimodal deformable registration~\cite{Qin2019Unsupervised}.
However, these works are for uni-modal or bi-modal data.
To our best knowledge, the potential of feature disentanglement for robust multimodal segmentation at arbitrary modality number has not been tapped yet. 

We propose a novel multimodal learning framework with feature disentanglement and gated feature fusion, which is robust to missing modalities.
Our network disentangles multimodal features into modality-specific appearance code and modality-invariant content code. The content code of each modality is fused into a shared representation containing discriminative information for segmentation task.
To enhance its modality-invariance, the shared representation is required to reconstruct any modality given corresponding appearance code, even in the absence of some modality(ies).
Furthermore, we employ a novel gated feature fusion strategy to automatically learn weight maps and gate the contribution from different modalities at different locations. 
We validate our proposed method on the task of multimodal brain tumor segmentation with BRATS challenge~\cite{menze2015multimodal}.
With competitive performance to the state-of-the-art methods for full modality,
our method is highly robust to various missing modality(ies) situations.

\section{Method}
\begin{figure*}
	\centering
	\includegraphics[width=1.0\textwidth]{{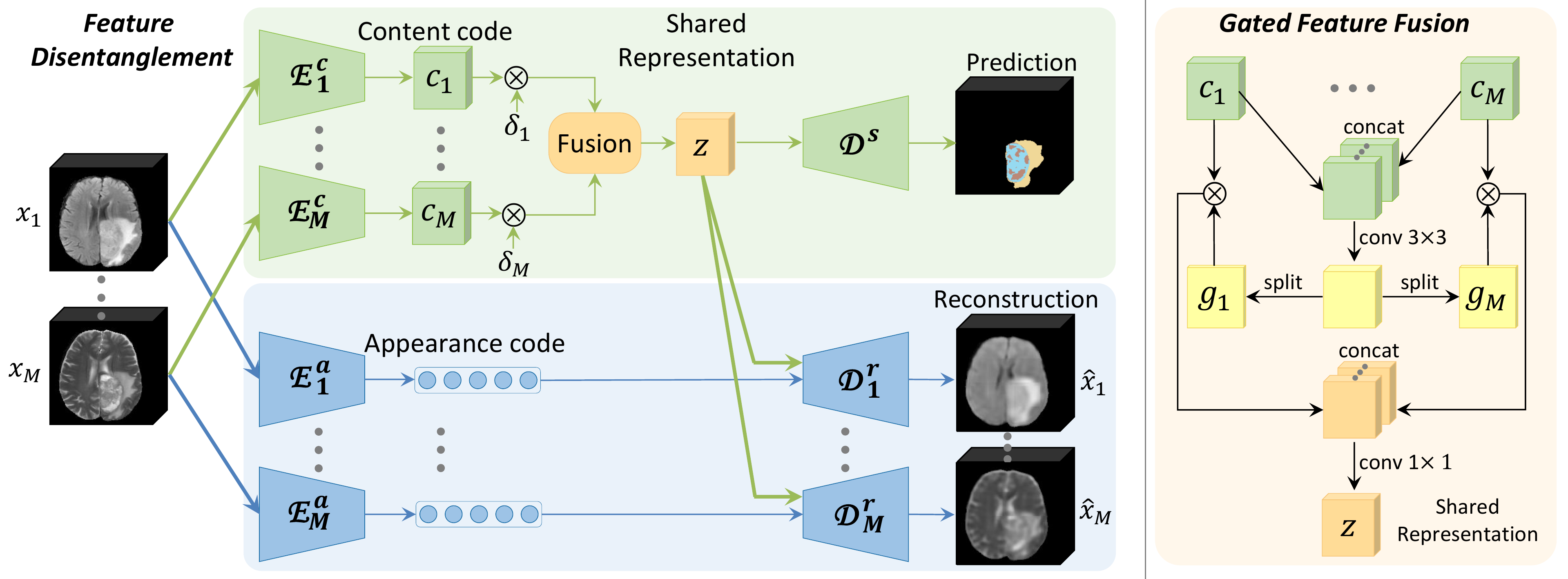}}
	\caption{Proposed multimodal segmentation framework. \textbf{Left}: Feature disentanglement for multimodal learning. \textbf{Right}: Detailed structure of the gated feature fusion module.}
\end{figure*}
An overview of our proposed multimodal segmentation framework is in Fig.~1.
We first introduce feature disentanglement to encode multimodal inputs to
modality-specific appearance code and modality-invariant content code.
Next, we present a learning-based gating strategy for integrating the complementary disentangled content code from individual modality to a more expressive fused representation, and detailed learning process and network architecture are described in the end.

\subsection{Feature Disentanglement for Robust Multimodal Learning}
We denote the multimodal images by $\{x_1,...,x_M\}$, where $M\!=\!4$ in our brain tumor segmentation task.
Each modality $x_i$ is input to its own appearance encoder $E_i^a$ and content encoder $E_i^c$, and we correspondingly obtain its disentangled appearance code $a_i = E_i^a(x_i)$ and content code $c_i = E_i^c(x_i)$.
For the appearance code, we follow the common practice in~\cite{lee2018diverse} and set it as a 8-bit vector assuming its prior distribution to be a centered isotropic Gaussian $\mathcal{N}(\textbf{0}, \textbf{\textit{I}})$.
The Kullback-Leibler (KL) divergence is computed to encourage the estimated distribution of $p(a_i)$ to be as close to the normal distribution. In this way, we obtain the loss of
$\mathcal{L}_{\text{KL}} \! = \! \sum_{i=1}^M \mathbb{E}[D_{\text{KL}}(p(a_i)||\mathcal{N}(\textbf{0},\textbf{\textit{I}}))]$ for training the appearance encoders $\{E_i^c\}$.

Next, for the content code $\{c_i\}$ which is expected to gain modality-invariance after evaporating the stylized appearances of different image modalities, we fuse them into an integrated representation $z \! = \! \mathcal{F} (\{c_i\})$ expressing essential semantic contents of the tumor.
$\mathcal{F}$ is an automatically learned fusion strategy for which we will elaborate in Section 2.2.
From the perspective of successful disentanglement, the obtained content representation $z$ should enable to be re-rendered into the original image given any appearance code of a certain modality.
To encourage such reconstruction capability, we develop the pseudo-cycle-consistency loss by introducing a set of modality-specific decoders $\{D_i^r\}$, as follows:
\begin{equation}
\mathcal{L}_{\text{rec}} = \sum_{i=1}^M||D_i^r(z,a_i)-x_i||_1, ~ \text{where} ~ z = \mathcal{F}(\delta_i c_1,...,\delta_M c_M),
\end{equation}
where we employ the L1-Norm to alleviate the generated images getting blurred.
With a Bernoulli indicator $\delta_i$, we are aiming to grant the content representation $z$ extra robustness to missing data, i.e., still producing a high-quality reconstructed $\hat{x}_i = D_i^r(z,a_i)$ even in the absence of $c_i$ when fusing content codes.
We perform the modality dropout in latent space by randomly setting $\delta_i$ to $0$, which turns off the content code $c_i$ in current learning iteration.

After the disentanglement procedure which cancels out the effect of modality-specific appearance features while aggregates the complementary content information from arbitrary combination of multimodal data, we can perform accurate and robust brain tumor segmentation.
We build a segmentation decoder $\hat{Y} \! = \! D^s(z)$ which learns discriminative pattern based on our derived representative and robust $z$. 
We jointly use Dice loss and weighted cross-entropy loss to handle the  unbalanced object sizes in multi-class segmentation:
\begin{equation}
\footnotesize
\mathcal{L}_{\text{seg}} \! = \! \textit{Dice}(\hat{Y}, Y) + H(\hat{Y}, Y)
= \! - \! \! \sum  \limits_{k \in K} \! \left(\frac{ 2 \sum_{j=1}^N y_j^k \hat{y}_j^k }{   \! \sum_{j=1}^N y_j^k y_j^k  + \hat{y}_j^k \hat{y}_j^k \! + \! \epsilon } + \sum \limits_{j=1}^{N}  w^{k} \! \cdot \! y_j^k \log q_j^k \right),
\end{equation}
where $y_j^k,q_j^k,\hat{y}_j^k$ respectively denote the ground truth, probability prediction and one-hot output of voxel $j$ for class $k$. Directly combing the two types of segmentation loss works well in practice, without need of particularly tuning a balancing weight between them.
The $\epsilon \! = \! e^{-7}$ is a constant for numerical stabilization, and $w^{k}$ is calculated online in a batch, treating class imbalance in cross-entropy loss.

\subsection{Multimodal Content Fusion with Learned Gating}
Effectively fusing the complementary information from various modalities is crucial in a multimodal learning framework. This also holds for our scenario, though we disentangle the content code and enforce robustness to missing data.
In fact, feature fusion plays a more important role in unusual inference situations such as unavailability of some modality(ies).
If not considered carefully, the fused representation would otherwise be infected by the noisy information from empty-input channel(s), then the model's performance is inevitably degraded.
Existing approaches tackle this using average~\cite{havaei2016hemis} or max operation~\cite{chartsias2018multimodal}.
However, the average operation makes each modality contribute equally, which may disregard highly informative features from a certain modality.
On the contrary, the max operation only retains the largest response, neglecting information from all the others.

Instead of hard-coding a fusion operation, we propose to automatically learn the mapping function to integrate multimodal features.
The contribution weights of a modality are not necessarily identical across locations, as a modality contains different amount of information for areas of each class. 
For example, T1c shows clear structure of enhancing tumor, but not for the area of edema.
In this regard, we dynamically learn a weight map to gate the scale of information from each content code $c_i$, with a flexibility of voxel-by-voxel.
Then, the gated content from individual modalities are fused to form the integrated representation.

Specifically, the disentangled content codes $\{c_1,...,c_M\}$ from each modality are concatenated, and then input to a convolutional layer with an output channel of $M$.
With sigmoid activation, we obtain the gating weight matrix $G$, which can be split into $M$ separate maps of $\{g_1,...,g_M\}$, one for each modality.
Next, we re-weight the content code as $\tilde{c}_i = c_i \cdot g_i$ with element-wise multiplication.
These outputs $\{\tilde{c}_i\}$ are concatenated and forwarded to a bottleneck $1\times1$ convolution, followed by Leaky ReLU activation.
As shown in Eq.~(1), we randomly set some content code(s) to be 0 with $\{\delta_i\}$ during training, to enhance model's robustness to missing data.
Overall, we obtain the fused content code $z = \mathcal{F}(\delta_i c_1,...,\delta_M c_M)$, which has the same feature map size and channel as individual code $c_i$.

It is worth noting that our learning-based gating strategy is general for multimodal feature fusion, which is superior to existing average or max hard-coding way, by properly aggregating complementary content with data-dependent weight.
In our framework, we jointly use it with the disentanglement procedure, and form an accurate and robust end-to-end multimodal learning method.

\subsection{Learning Process and Network Architecture}
The entire framework is learned with the overall objective function as:
\begin{equation}
\mathcal{L}_{\text{total}} (D^s, \{E_i^c, E_i^a, D_i^r\})
=  \mathcal{L}_{\text{seg}}+\lambda \mathcal{L}_{\text{rec}}+\beta \mathcal{L}_{\text{KL}},
\end{equation}
where the $\lambda,\beta$ are trade-off parameters weighting the importance of each component, which are both empirically set as $0.1$ in our experiments.
We utilize an Adam optimizer with an initial learning rate of $1e^{-4}$, and progressively multiply it by $(1-\text{epoch}/\text{max\_epoch})^{0.9}$ during training.
The intensive components with our model only allow us to set the batch size as 1 using one Nvidia Xp GPU. 

Our encoders $\{E_i^c\}$ and decoder $D^s$ for segmentation task adopt 3D U-Net~\cite{cciccek20163d} architecture, except that one independent encoder is used for each input modality. In each downsampling stage, the content features of individual modality are fused via the learning-based gating strategy with 0.5 probability of $\delta_i$ to be zero for dropping $c_i$.
Each fused features are then skip-connected to the corresponding upsampling stage.
Each $E_i^c$ consists of 4 residual blocks with instance normalization and Leaky ReLU activation. 
Between each block, the image dimensions are progressively reduced by 2 and the feature channels are doubled. All convolutions use kernel size of $3\! \times \! 3 \! \times \! 3$ and the initial channel number is 16. The $D^s$ also has 4 residual blocks which is similar to $E_i^c$, except that the feature map size is upsampled by 2 with channel number halved after each stage. 
For image reconstruction, we generally follow the practice in~\cite{huang2018multimodal}.
Specifically, each $E_i^a$ consists of 5 convolutional layers followed by a global average pooling and a fully-connected layer to obtain the appearance code. 
Each $D_i^r$ uses 4 residual blocks followed by 4 upsampling and convolutional layers to produce $\hat{x}_i$.

\section{Experiments}
\textbf{Dataset and Preprocessing.}
We validate our proposed method with the 2015 Brain Tumor Segmentation Challenge (BRATS) dataset~\cite{menze2015multimodal}. 
The training set consists of 274 cases with ground truth being provided. The test set contains
\begin{table*}
	\caption{Comparison of brain tumor segmentation performance on BRATS 2015 test set. The values are obtained by submitting our results to the online evaluation system. 
	}
	\centering
	\begin{center}
		\resizebox{0.9\textwidth}{!}{%
			\begin{tabular}{l|ccc|ccc|ccc}
				\hline
				
				\multirow{2}{*}{Methods} &\multicolumn{3}{c|}{Dice(\%)} &\multicolumn{3}{c|}{Precision(\%)} &\multicolumn{3}{c}{Sensitivity(\%)}\\
				\cline{2-10}
				{}&Complete &Core &Enhancing &Complete &Core &Enhancing &Complete &Core &Enhancing\\
				
				\hline
				Kamnitsas et al.~\cite{kamnitsas2017efficient} &84 &67 &63 &82 &\textbf{85} &\textbf{64} &\textbf{89} &62 &66\\
				
				\hline
				Zhao et al.~\cite{zhao2018deep} &82 &\textbf{72} &62 &84 &78 &60 &83 &\textbf{73} &69\\
				
				\hline
				OM-Net~\cite{zhou2018one} &\textbf{86} &71 &\textbf{64} &\textbf{86} &83 &61 &88 &68 &\textbf{72}\\
				
				\hline
				
				Ours &84 &\textbf{72} &\textbf{64} &84 &80 &\textbf{64} &\textbf{89} &69 &68\\
				
				\hline
		\end{tabular}}
	\end{center}
\end{table*}
\begin{table*}[h!]
	\caption{Robustness comparison of our method against HeMIS~\cite{havaei2016hemis} and the imputation MLP~\cite{havaei2016hemis} on the test split of BRATS 2015 training set. The Dice score is presented for every combination case of modalities being available ($\checkmark$) or missing ($-$).
	}
	\centering
	\begin{center}
		\resizebox{0.9\textwidth}{!}{%
			\begin{tabular}{cccc|ccc|ccc|ccc}
				\hline
				
				\multicolumn{4}{c|}{\multirow{2}{*}{Modalities}} &\multicolumn{9}{c}{Dice(\%)}\\
				
				\cline{5-13}
				\multicolumn{4}{c|}{}&\multicolumn{3}{c|}{Complete} &\multicolumn{3}{c|}{Core} &\multicolumn{3}{c}{Enhancing}\\
				
				\hline
				
				F &T1 &T1c &T2 &~~Ours &~~HeMIS~ &~~MLP~~ &~~Ours &~~HeMIS &~~MLP~~ &~~Ours &~~HeMIS &~~MLP~~\\
				\hline
				--  &--  &--  &\checkmark &~~\textbf{85.49} &~~58.48 &~~61.50~~ &~~\textbf{58.66} &~~40.18 &~~37.32~~ &~~\textbf{37.66} &~~20.31 &~~18.62~~\\
				--  &--  &\checkmark  &-- &~~\textbf{71.86} &~~33.46 &~~2.04~~ &~~\textbf{72.87} &~~44.55 &~~17.70~~ &~~\textbf{70.22} &~~49.93 &~~32.92~~\\
				--  &\checkmark  &--  &-- &~~\textbf{68.40} &~~33.22 &~~2.07~~ &~~\textbf{50.00} &~~17.42 &~~10.52~~ &~~\textbf{22.67} &~~4.67 &~~10.78~~\\
				\checkmark  &--  &--  &-- &~~\textbf{83.02} &~~71.26 &~~63.81~~ &~~\textbf{46.67} &~~37.45 &~~34.26~~ &~~\textbf{28.30} &~~5.57 &~~15.90~~\\
				--  &--  &\checkmark  &\checkmark &~~\textbf{87.53} &~~67.59 &~~64.97~~ &~~\textbf{78.46} &~~63.39 &~~49.38~~ &~~\textbf{76.82} &~~65.38 &~~60.30~~\\
				--  &\checkmark  &\checkmark  &-- &~~\textbf{74.59} &~~45.93 &~~1.99~~ &~~\textbf{76.40} &~~55.06 &~~26.55~~ &~~\textbf{73.95} &~~62.40 &~~40.93~~\\
				\checkmark  &\checkmark  &--  &-- &~~\textbf{87.66} &~~80.28 &~~78.13~~ &~~\textbf{60.17} &~~49.52 &~~48.97~~ &~~\textbf{35.28} &~~22.26 &~~25.18~~\\
				--  &\checkmark  &--  &\checkmark &~~\textbf{87.87} &~~69.56 &~~66.88~~ &~~\textbf{64.88} &~~47.26 &~~43.66~~ &~~\textbf{41.05} &~~23.56 &~~26.37~~\\
				\checkmark  &--  &--  &\checkmark &~~\textbf{89.08} &~~82.10 &~~81.35~~ &~~\textbf{63.51} &~~53.42 &~~52.41~~ &~~\textbf{39.72} &~~23.19 &~~25.01~~\\
				\checkmark  &--  &\checkmark  &-- &~~\textbf{88.01} &~~79.80 &~~81.13~~ &~~\textbf{78.09} &~~66.12 &~~65.51~~ &~~\textbf{76.62} &~~67.12 &~~66.19~~\\
				\checkmark  &\checkmark  &\checkmark  &-- &~~\textbf{87.73} &~~80.88 &~~82.19~~ &~~\textbf{80.68} &~~69.26 &~~69.34~~ &~~\textbf{78.81} &~~71.30 &~~70.93~~\\
				\checkmark  &\checkmark  &--  &\checkmark &~~\textbf{89.07} &~~83.87 &~~80.40~~ &~~\textbf{65.99} &~~57.76 &~~53.46~~ &~~\textbf{43.04} &~~28.46 &~~28.34~~\\
				\checkmark  &--  &\checkmark  &\checkmark &~~\textbf{89.06} &~~82.78 &~~83.37~~ &~~\textbf{79.47} &~~70.62 &~~70.45~~ &~~\textbf{78.07} &~~70.52 &~~70.56~~\\
				--  &\checkmark  &\checkmark  &\checkmark &~~\textbf{88.26} &~~70.98 &~~67.85~~ &~~\textbf{80.84} &~~66.60 &~~55.40~~ &~~\textbf{78.56} &~~67.84 &~~64.81~~\\
				\checkmark  &\checkmark  &\checkmark  &\checkmark &~~\textbf{89.07} &~~83.15 &~~82.43~~ &~~\textbf{81.19} &~~72.50 &~~71.46~~ &~\textbf{~79.13} &~~75.37 &~~72.08~~\\
				\hline
				\multicolumn{4}{c|}{Average} &~~\textbf{84.45} &~~68.22 &~~60.01~~ &~~\textbf{69.19} &~~54.07 &~~47.09~~ &~~\textbf{57.33} &~~43.86 &~~41.93~~\\
				
				\hline
		\end{tabular}}
	\end{center}
\end{table*}110 cases with reference labels being held by the organizers and the evaluation can be obtained via an online system.
Each case contains four MRI modalities including FLAIR, T1, T1c, and T2.
The task of the challenge is to segment three tumor classes, i.e. complete, core, and enhancing tumor.
The dataset is preprocessed with being skull-stripped, co-registered, and resampled to isotropic $1 mm^3$ resolution, by the organizer. We further normalized the intensity of each volume to zero mean and unit variance within the brain tissue area.
A patch of size $80\times80\times80$ was randomly cropped during training as input to the network. 
\\
\\
\textbf{Performance of Robust Brain Tumor Segmentation.}
		We first compare our method with the state-of-the-art methods on the test set of BRATS 2015 for full modalities. 
		Results were obtained directly from the online evaluation system and compared without postprocessing.
		In Table~1, our method achieves the highest Dice score of the core and enhancing tumor, with the other evaluations highly competitive with the ranking 1st approach OM-Net~\cite{zhou2018one}, validating the effectiveness of our segmentation backbone.
		
		\begin{figure*}
			\centering
			\includegraphics[width=0.88\textwidth]{{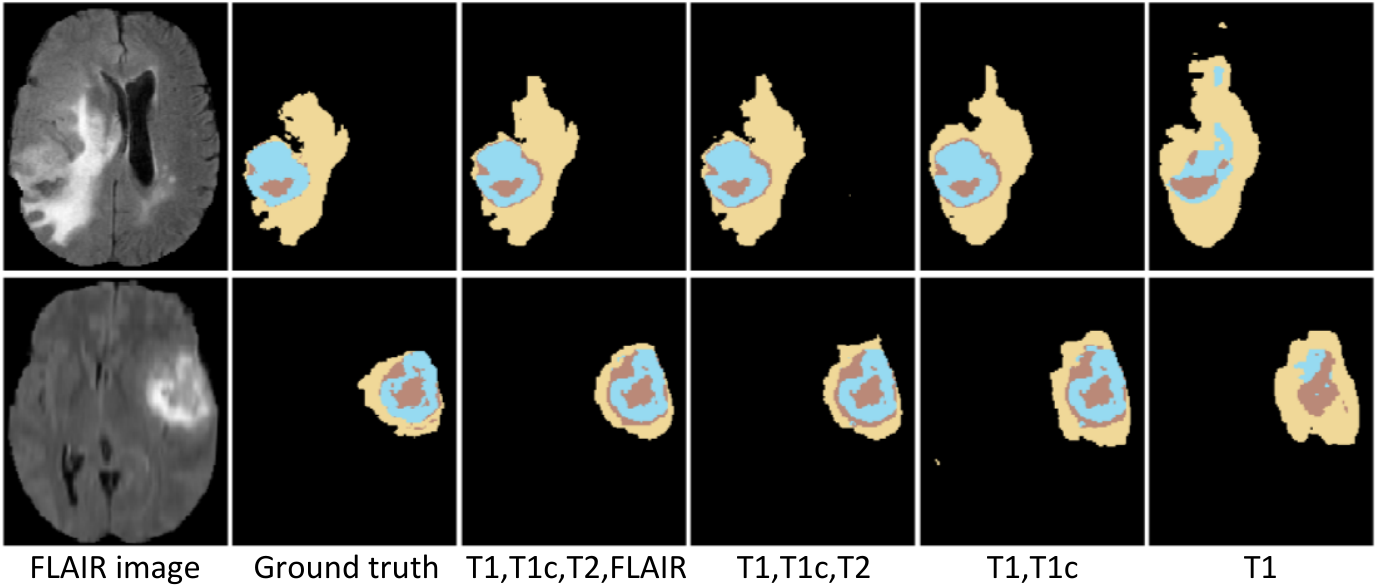}}
			\caption{Segmentation results from our method for complete tumor (yellow), tumor core (red), and enhancing tumor (blue). Input modalities at inference are indicated.}
		\end{figure*}
		We then evaluate the robustness of our method to missing modality inference. The absence of modality $i$ is implemented by setting $\delta_i$ to be zero for dropping $c_i$ at inference. For direct comparison with the HeMIS method and image synthesis method using multilayer perceptron (MLP)~\cite{havaei2016hemis}, we used the same data split of BRATS training set as in \cite{havaei2016hemis} and directly referenced the results from their paper. 
		In Table 2, our method significantly outperforms the HeMIS and imputation MLP methods for all the 15 possible combination situations of unavailable modalities and all the three tumor classes. This demonstrates the outstanding robustness of our multimodal segmentation method.
		From the results, we can see that FLAIR and T2 modalities 
		are more informative than others
		for the complete tumor segmentation, and T1c is discriminative for accurate prediction of enhancing tumor. 
		In Fig.~2, we show that with the increase of the number of missing modalities, the segmentation results produced by our robust model just gradually degrade, rather than encountering sudden failure.
		Even with T1 alone, we can achieve decent segmentation for the complete tumor and tumor core. 
\\
\\
\textbf{Ablation Study.}
		We investigate the effectiveness of feature disentanglement and gated fusion, as two key components in our method.
		We first set up a baseline network which uses average fusion without feature disentanglement. Then we add the feature disentanglement and gated fusion one by one into the baseline network. In Fig.~3~(a), we compare the performance of the three networks on the Dice score, averaging over the 15 possible combination situations of input modalities. 
		Both the feature disentanglement and gated fusion bring performance improvement across all the tumor parts, achieving the highest Dice score in most situations (10, 13, and 11 situations out of 15 for the complete tumor, tumor core, and enhancing tumor respectively). 
		Fig.~3~(b) shows the reconstruction results of FLAIR and T2 image by combining their corresponding appearance code with the shared representation fused from content code of different combination of input modalities.
		\begin{figure*}
			\centering
			\includegraphics[width=0.92\textwidth]{{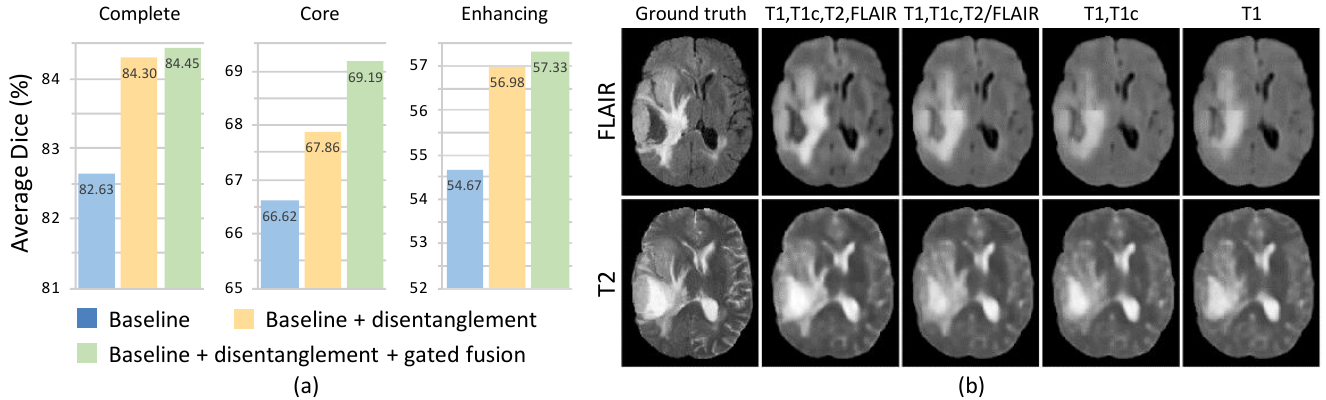}}
			\caption{(a) Ablation study of key components in our method. (b) Example reconstruction of FLAIR and T2 images for different combinations of input modalities.
			}
		\end{figure*}
		Even when some modality(ies) are missing, our network can still reconstruct the missing modality with the shared representation, indicating that the shared representation $z$ successfully yields the essential tumor content.

\section{Conclusion}
We propose a novel multimodal segmentation framework 
which jointly uses the feature disentanglement and gated feature fusion to obtain a modality-invariant and discriminative representation. We validate our method on brain tumor segmentation under both full modalities and various combination situations of missing modalities, achieving new state-of-the-art results on BRATS benchmark.
The outstanding robustness to great inference variations can make our method widely-applicable in real-world clinical scenarios.
\\
\\
\textbf{Acknowledgments.} This work was supported in part by the National Basic Program of China 973 Program under Grant 2015CB351706, the National Natural Science Foundation of China, under Project No. U1613219, the Research Grants Council of Hong Kong Special Administrative Region, under Project No. CUHK14225616, and the Hong Kong Innovation and Technology Commission, under Project No. ITS/319/17.

\bibliographystyle{splncs04.bst}

\bibliography{reference}

\end{document}